\documentclass[10pt,twocolumn,letterpaper]{article}

\usepackage[cvpr]{cvpr}      

\usepackage{graphicx}
\usepackage{amsmath}
\usepackage{amssymb}
\usepackage{booktabs}
\graphicspath{./images/}

\usepackage[pagebackref,breaklinks,colorlinks]{hyperref}

\usepackage[capitalize]{cleveref}
\crefname{section}{Sec.}{Secs.}
\Crefname{section}{Section}{Sections}
\Crefname{table}{Table}{Tables}
\crefname{table}{Tab.}{Tabs.}

\def\cvprPaperID{15} 
\def\confName{CVPR}
\def\confYear{2022}

\begin{document}

\title{Light Weight Character and Shape Recognition for Autonomous Drones}

\author{Neetigya Poddar \\
Manipal Institute of Technology\\
Eshwar Nagar, Manipal\\
{\tt\small neetigyapoddar1@gmail.com}
\and
Shruti Jain\\
Manipal Institute of Technology\\
Eshwar Nagar, Manipal\\
{\tt\small shrutipraveenjain@gmail.com}
}
\maketitle

\begin{abstract}
    There has been an extensive use of Unmanned Aerial Vehicles in search and rescue missions to distribute first aid kits and food packets. It is important that these UAVs are able to identify and distinguish the markers from one another for effective distribution. One of the common ways to mark the locations is via the use of characters superimposed on shapes of various colors which gives rise to wide variety of markers based on combination of different shapes, characters, and their respective colors. 
   In this paper, we propose an object detection and classification pipeline which prevents false positives and minimizes misclassification of alphanumeric characters and shapes in aerial images. Our method makes use of traditional computer vision techniques and unsupervised machine learning methods for identifying region proposals, segmenting the image targets and removing false positives. We make use of a computationally light model for classification, making it easy  to be deployed on any aerial vehicle. 
\end{abstract}

\section{Introduction}
\label{sec:intro}
The use of drones for distributing food packets and first aid kits in search and rescue missions has increased in the recent years due to its versatility and fast speed compared to a human alternative. By making use of characters superimposed on shapes, we can mark and identify many unique locations identified by the character, shape and their colors. Associations like AUVSI (Association for Unmanned Vehicle Systems International) have taken a keen interest in this use case for UAVs and have organized the AUVSI-SUAS competition around it. Our robotics team 
participated in this competition and this paper is based on our approach for solving the tasks presented in the rule book \footnote{link to the \href{https://static1.squarespace.com/static/5d554e14aaa5e300011a4844/t/619908c51e400057a8a791e6/1637419206568/auvsi_suas-2022-rules.pdf}{rule book}}. 
One of the main challenges of using drones to distribute essential kits in search and rescue missions is that these drones need to recognize the drop off and collection locations precisely from a height of over 150 ft. The characters and shape will also have an arbitrary angle with respect to the drone, adding another challenge for the classifier. 

Most of the methods existing for object detection and classification from aerial vehicles make use of computationally heavy deep learning methods depending on a large labelled dataset making it unfeasible to deploy. Our approach first detects areas of interest that are then segmented into binary masks containing either the shape, letter or the background. After removing the false positives and the background masks, the remaining masks are passed to the classifier which classifies the character, the shape and their respective colors. We make use of traditional image processing techniques to make it robust to noise and lighting changes while keeping it computationally light.
\\


\begin{figure}[h]
    \centering
    \includegraphics[scale=1]{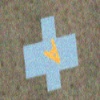}
    \caption{An example of an image target.}
    \label{fig: target_example}
\end{figure}


\section{Related Work}


Object detection and classification has been extensively explored in the recent decades. We have split this section in the following sub sections (1) Aerial object Detection  (2) Unsupervised Image Segmentation. Most of the approaches use deep learning based methods which are summarized in this section.

\subsection{Aerial object Detection}
Object Detection in Aerial images is more challenging and demands separate attention because targets are small and sparse with concentrations over minority regions. 

Semantic segmentation and detection is integrated in \cite{segment_before_classify} in order to improve performance.  Fast R-CNN \cite{fast_rcnn} is extended in \cite{fast_vehicle} for detecting vehicles in aerial images. Paper \cite{roi_transformer} investigates misalignment between ROI and objects in aerial image detection. It introduces a ROI transformer to address this issue. A scale adaptive proposal network is also proposed for object detection in aerial images in \cite{scale_adaptive}.

\subsection{Unsupervised Image Segmentation} 
The majority of unsupervised image segmentation methods use local patch cues like colour, brightness, or texture to create pixel-level grouping. Among these schemes, the three most widely-used methods include Shi and Malik’s Normalized Cuts \cite{spectral_segmentation} \cite{normalizedcuts_segmentation} and Felzenszwalb and Huttenlocher’s graph-based method \cite{graphbased_segmentation}. A unified approach for bottom-up multi-scale hierarchical image segmentation is proposed in \cite{DBLP}.


\section{Problem Formulation}
The objective of the AUVSI-SUAS task is to detect and classify the characters, shape, and their respective colors. In accordance with the rules of AUVSI-SUAS 2022, the shapes that we consider are square, rectangle, triangle, trapezoid, hexagon, heptagon, octagon, quarter circle, semi-circle, stars, crosses, circle and pentagon.
The alphanumeric characters are all the numbers and alphabets in the English language.
We have White, Black, Gray, Red, Blue, Green, Yellow, Purple, Brown and Orange as the possible classes for color classification. 
\\
The term 'false positives' in our paper indicates the background instances that are incorrectly classified as targets.
The term 'masks' refers to binary thresholded gray scaled images that have been segmented out from the images.

\section{Methodology}
The pipeline has 4 integral steps : Region of Interest Detector, Segmentation, False Positive Removal System and Classification.
These four parts work in tandem to provide results as accurately as possible. 

\begin{figure}[h]
    \centering
    \includegraphics[scale=0.55]{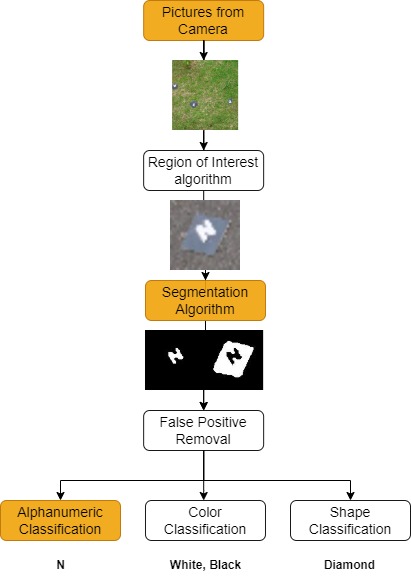}
    \caption{Our pipeline for Object detection and classification}
    \label{fig: pipeline}
\end{figure}

\subsection{Region of Interest Detector}
\subsubsection{Pre-Processing}
 The pictures from the camera on the drone are sent to the Region of Interest Detector which finds the possible areas in the picture that could contain the characters and the shapes. The raw picture is passed to the pre-processing module that reduces the colors in the image, thus smoothening it, by using K-Means clustering with a high k value. This smoothened picture is converted to Hue Saturation Value (HSV) color space and denoised by OpenCV functions. 
 \subsubsection{Getting Bounding boxes}
 To get the bounding boxes around the possible regions of interest,  we first pass the pre-processed image through canny edge detection. After getting the edges, we make the bounding boxes around the contours. This usually returns a large amount of bounding boxes and we filter them out by putting constraints on the bounding box size. To suppress the redundant bounding boxes, we discard those which have a high Intersection over Union (IOU) with the other boxes. We iterate this process a number of times till we are certain that all redundant boxes have been eliminated.   
 
\subsection{Segmentation}

Segmentation helps the model to focus on the important parts of the picture. In our segmentation, we try to get rid of the different colors and separate the character and the shape from each other as well as the background. This helps the model to not over-fit to the background as well as the relative positioning of the character to the shape.

\begin{figure}[h]
    \centering
    \includegraphics[scale=0.4]{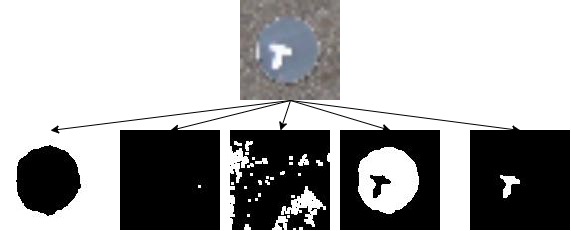}
    \caption{An example of segmentation on an image target}
    \label{fig: segmentation}
\end{figure}

\subsubsection{Preprocessing}
The picture first undergoes a pre-processing stage where Principal Component Analysis is used on the image in the Hue Lightness Saturation (HLS) color space by preserving 95\% variance in the picture. 
Then we scale them using Min-Max Scaler to reduce the processing time. We present the preprocessed image to K-Means clustering algorithm. 

\subsubsection{Mask creation}
We segment the image into multiple masks based on the pixel values of the image using K-means clustering algorithm with a k value of 5. Out of the 5 masks, one of them contains the letter mask and one of them the shape mask. The remaining three masks are discarded by the false positive removal module.
\subsubsection{Reconstruction of centers of clusters}
The center of the clusters returned by the algorithm is calculated by taking the average of the cluster points, which is very sensitive to outliers. For this reason, we use median of the cluster points, which improved the accuracy of color classification considerably. To get the HLS pixel values from the center of the clusters, we inverse the scaled PCA features. These pixel values are further passed to the color classification module to get the color names. 
Each individual cluster constitutes a possible important mask so we reconstruct these clusters into a binary mask with the foreground in white and use this for classification.

\subsection{False Positive Removal}

Since we are using traditional image processing for ROI Detection and unsupervised learning for segmentation, there is a possibility for false positives in our images. Reducing False Positives is important because it is very costly if the drone keeps on arriving at the wrong locations.  

\begin{figure}[h]
    \centering
    \includegraphics[scale=0.35]{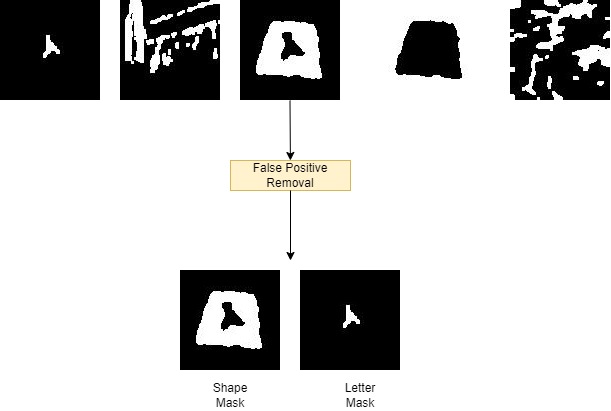}
    \caption{An example of False Positive Removal system}
    \label{fig: false positive}
\end{figure}

We did false positive removal in two stages:

\subsubsection{Counting the Regions}
We count the number of regions present in the mask. Two regions are said to be connected if they have one or more pixels connecting them either vertically, horizontally or diagonally. If the number of connected regions are more than a threshold value, we discard the mask as noise. 
The idea behind this is that noise has a high entropy and would be spread all over the mask, but the foreground will be concentrated over a smaller area and should ideally have just one connected region. Our experiments have showed that a threshold value of 2 works the best. Most of the false positives were eliminated by this stage itself.

\subsubsection{Classification Probability Thresholding}
We analyze the classification model's probability distribution on the input masks. If the probability for each of the classes is below a certain threshold then we assume that the input is out of distribution and is probably noise and is discarded.
The result of False Positive Removal is very promising and was able to compensate for the simplistic algorithms that we used in the start.
 
\subsection{Classification}
\begin{figure}[h]
    \centering
    \includegraphics[scale=0.5]{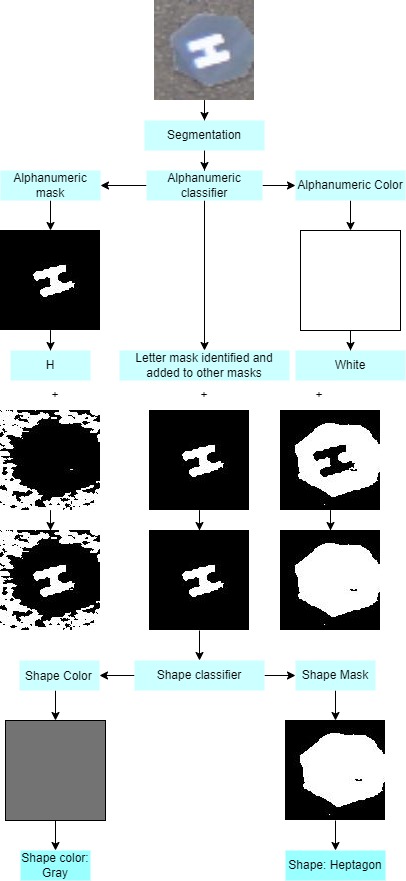}
    \caption{Classification System Pipeline.}
    \label{fig: classification}
\end{figure}

The masks are received from the segmentation module after being filtered from the false positive removal system. These masks are passed first to the alphanumeric classification. Based on the stage two of the false positive removal system, only the correct alphanumeric mask with its correct class remains. We then superimpose this mask on all the other masks so that the empty space due to the character in the correct shape mask gets filled. These masks are then passed to the shape classifier. We get the correct shape mask along with its class as the output from the classifier. Now, the shape mask and the character mask are passed to the color classifier to get their respective colors. The pipeline is explained more clearly in \ref{fig: classification}.

\subsubsection{Alphanumeric Classification} 
Dataset: The model was trained on binary masks from LMNIST, char74k dataset and previous year AUVSI-SUAS dataset on Kaggle \footnote{link to the AUVSI-SUAS dataset:  \href{https://www.kaggle.com/gndctrl2mjrtm/auvsi-suas-dataset}{https://www.kaggle.com/gndctrl2mjrtm/auvsi-suas-dataset}}. We augmented the dataset by rotating it, performing horizontal shift, vertical shift and a zoom range of 0.2. 

The model used for the classification is Xception Net \cite{xception} which was pre-trained on ImageNet weights \cite{imagenet}. We have used Batch Normalization and Dropout to overcome overfitting.  

\subsubsection{Shape Classification} 
Dataset: The dataset was formed by synthesising the masks for different shapes and handpicking the masks that we got after passing the AUVSI-SUAS dataset to the segmentation module. This dataset was further augmented by rotating it, flipping it, performing vertical and horizontal shifts.

The shape classifier was trained in a similar manner to the alphanumeric classifier using an Xception Network \cite{xception}, pretrained on imagenet weights \cite{xception}. 


\subsubsection{Color Classification}
After the masks are passed through the above classifiers, those capturing background noise are removed and hence the remaining masks only contain either the letter or the shape. The cluster centers of these masks, as given by the segmentation module are used to classify the colors. The HLS value of the cluster center was converted to Red Green Blue(RGB) color space and compared with the RGB values of standard colors. We chose Manhattan distance as a measure of similarity between the RGB values.


\section{Experiments}
During the development process, we theorized, implemented and analysed a number of different methods before the above mentioned pipeline was finalised.  
\subsection{End to End Deep Learning}
\subsubsection{Procedure}
Another approach to solving the problem would be training an end to end Convoluted Neural Network (CNN) to classify the character, shape and color. The input to this Convoluted neural network would be the cropped images from RoI detector.
\subsubsection{Problems faced}
One of the major problems with this approach is acquiring the dataset for a unique problem like this. There are a lot of permutations possible of each shape with every character angled at different orientations. Additionally, the letters, shapes and backgrounds can be of different types with different colors. This coupled with the fact that such a dataset does not currently exist and needs to be created by us, makes us believe that the model would overfit and not generalize enough for observed data. The only dataset that was the closest to what we needed was the previous year AUVSI-SUAS dataset, but we could not use it because it did not contain numbers and it did not have all shapes as specified in the AUVSI-SUAS 2022 rulebook.

\subsection{Neural Net for False Positives}
\subsubsection{Procedure}
A Neural Network as a False Positive Removal system can be trained to reject the background masks instead of the traditional image processing methods. An alternative network structure was also thought of where a model will classify whether the mask is letter, shape or noise. Training for the noise class can be done using Negative sampling where pictures with random pixels are filled.
\subsubsection{Problems Faced}
Training such a model would be difficult with a chance of the model overfitting on the noise presented to it and the dataset required would be difficult to acquire and not trustworthy.

\subsection{Color Spaces for Pre-Processing}
We experimented with different color spaces like HSL, HSV, RGB, and LAB. We found color spaces different from RGB giving better results as they were more uniform and had clear distinctions between the colors. Ultimately, we chose HSL and HSV as our primary color spaces.

\subsection{Different Algorithms for Segmentation}
We experimented with DBSCAN Clustering, Watershed algorithm and Gaussian Mixture.
\subsubsection{DBSCAN Clustering}
We could not fix the number of clusters in this algorithm. This led to the character and the shape masks being broken as the algorithm failed to see some points dense enough to consider it a single cluster. 
\subsubsection{Watershed Algorithm}
Watershed algorithm is quite dependent on the edges of the image that are provided to it.
In our case, the character boundary can touch the shape boundary, causing problems for the filling algorithm.
Additionally, the images that we have, have been taken from a height of 150 feet and the image targets have been zoomed in a lot to get the closely cropped images from the ROI Detector. This results in noisy images, due to which the edges were not sharp enough at times to give distinct boundaries to the algorithm.

\subsubsection{Gaussian Mixtures}
Gaussian Mixture gave promising results but setting the hyperparameters was quite tedious and it did not give any considerable improvement as compared to K-Means clustering.
We experimented with a custom version of DBScan: We appended the (x,y) coordinates along with the RGB values of the pixels and fed it into Gaussian Mixtures and it gave better results initially. However, it also gave bloated and inaccurate masks for characters. It clustered together pixels which were of different color but still close together which resulted in bloating of masks. 

\subsection{Other Neural Net architectures for classification:}
Upon testing initially, we did not get great results from CNN and we tried to explore better and more robust algorithms. We tried alternatives like Convex Hull, Siamese Network, Scale Invariant Feature Transform approaches to improve generalization. Convex Hulls did not work out because a lot of shapes are not convex polygons (example: stars, crosses) so it failed in those scenarios. Siamese Networks turned out to be quite dependent on the scaling and the relative positioning of the mask in the photograph. That is why we improved the dataset generator and got better results in the CNN.

\section{Results}

\begin{center}
    \includegraphics[scale=0.55]{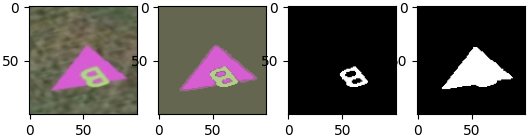}

    \includegraphics[scale=0.55]{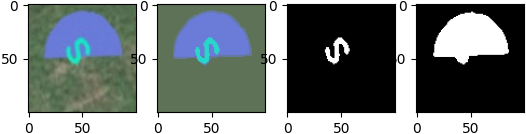}
    \captionof{figure}{Example of Results}
\end{center}

We tested our approach on pictures taken from a rooftop of a 12-storey building which is around 150 feet. The results of the testing are summarized below. 
\begin{table}[h!]
    \label{tab:table1}
    \centering
    \begin{tabular}{|c|c|}
    \hline
         \textbf{Field in which result was observed} & \textbf{Value observed} \\ 
         \hline
         Percentage of detected targets & 55\% \\
         \hline
         Alphanumeric Character Accuracy &\\ in detected targets & 75\% \\
         \hline
         Shape Accuracy in detected targets & 55\% \\
         \hline
         False Positives & 14\% \\
         \hline
    \end{tabular}
\end{table}

Considering that our model was tested on completely out of distribution data, the results seem promising. One of the major problems faced by the character classifier was that some rotated characters resembled some other characters. For example, a rotated 'N' looks like a 'Z'. The shape classifier performed poorly on shapes like hexagon, heptagon, octagon and circle because the masks were not precise and the edges were blurred, making these shapes indistinguishable.
The ROI Detector captures all of the image targets. The percentage of targets that were detected is low because the image targets get filtered out in the second stage of False Positive Removal module. This is primarily because since the data that it is tested on is out of distribution, the probability distribution of the classifiers are more flattened, discarding some image targets as noise. We can increase the number of image targets detected at the cost of increasing the percentage of false positives. 


\section{Conclusion}

Recent years have seen a surge in the use of aerial autonomous drones. This paper aims to help in identifying various landing and drop-off locations for the drones to which they can deliver without any manual input. This pipeline has been deployed on our drone with which we participated in the AUVSI-SUAS Competition. 
The computational inexpensiveness of our pipeline helps to keep the cost low and such technology to be affordable even in cheap, light drones. It is a step forward in the cognification of aerial vehicles.

{\small
\bibliographystyle{ieee_fullname}
\bibliography{LightWeightShapeClassifier}
}

\end{document}